# Compositional Generalization in Multilingual Semantic Parsing over Wikidata

**Ruixiang Cui, Rahul Aralikatte, Heather Lent** and **Daniel Hershcovich**
Department of Computer Science
University of Copenhagen
`{rc, rahul, hcl, dh}@di.ku.dk`

## Abstract

Semantic parsing (SP) allows humans to leverage vast knowledge resources through natural interaction. However, parsers are mostly designed for and evaluated on English resources, such as CFQ (Keysers et al., 2020), the current standard benchmark based on English data generated from grammar rules and oriented towards Freebase, an outdated knowledge base. We propose a method for creating a multilingual, parallel dataset of question-query pairs, grounded in Wikidata. We introduce such a dataset, which we call Multilingual Compositional Wikidata Questions (MCWQ), and use it to analyze the compositional generalization of semantic parsers in Hebrew, Kannada, Chinese and English. While within-language generalization is comparable across languages, experiments on zero-shot cross-lingual transfer demonstrate that cross-lingual compositional generalization fails, even with state-of-the-art pretrained multilingual encoders. Furthermore, our methodology, dataset and results will facilitate future research on SP in more realistic and diverse settings than has been possible with existing resources.

## 1 Introduction

Semantic parsers grounded in knowledge bases (KBs) enable knowledge base question answering (KBQA) for complex questions. Many semantic parsers are grounded in KBs such as Freebase (Bollacker et al., 2008), DBpedia (Lehmann et al., 2015) and Wikidata (Pellissier Tanon et al., 2016), and models can learn to answer questions about unseen entities and properties (Herzig and Berant, 2017; Cheng and Lapata, 2018; Shen et al., 2019; Sas et al., 2020). An important desired ability is compositional generalization—the ability to generalize to unseen *combinations* of known components (Oren et al., 2020; Kim and Linzen, 2020).

One of the most widely used datasets for measuring compositional generalization in KBQA is CFQ

| Lang. | Question |
|---|---|
| En | Did Lohengrin 's male actor marry Margarete Joswig |
| He | האם ה שחקן ה גברי של לוהנגרין התחתן עם מרגרט יוסוויג |
| Kn | ಲೋಹೆಂಗ್ರಿನ್ ಅವರ ಪುರುಷ ನಟ ವಿವಾಹವಾದರು ಮಾರ್ಗರೇಟ್ ಜೋಸ್ವಿಗ್ |
| Zh | Lohengrin 的 男 演员 嫁给 了 Margarete Joswig 吗 |

SPARQL Query:
`ASK WHERE { ?x0 wdt:P453 wd:Q50807639 . ?x0 wdt:P21 wd:Q6581097 . ?x0 wdt:P26 wd:Q1560129 . FILTER ( ?x0 != wd:Q1560129 )}`

Figure 1: An example from the MCWQ dataset. The question in every language corresponds to the same Wikidata SPARQL query, which, upon execution, returns the answer (which is positive in this case).

(Compositional Freebase Questions; Keysers et al., 2020), which was generated using grammar rules, and is based on Freebase, an outdated and unmaintained English-only KB. While the need to expand language technology to many languages is widely acknowledged (Joshi et al., 2020), the lack of a benchmark for compositional generalization in multilingual semantic parsing (SP) hinders KBQA in languages other than English. Furthermore, progress in both SP and KB necessitates that benchmarks can be reused and adapted for future methods.

Wikidata is a multilingual KB, with entity and property labels in a multitude of languages. It has grown continuously over the years and is an important complement to Wikipedia. Much effort has been made to migrate Freebase data to Wikidata (Pellissier Tanon et al., 2016; Diefenbach et al., 2017; Hogan et al., 2021) but only in English. Investigating compositional generalization in cross-lingual SP requires a multilingual dataset, a gap we address in this work.

We leverage Wikidata and CFQ to create **M**ultilingual **C**ompositional **W**ikidata **Q**uestions (MCWQ), a new multilingual dataset of compositional questions grounded in Wikidata. Beyond the original English, an Indo-European language using the Latin script, we create parallel datasets of questions in Hebrew, Kannada and Chinese, which use different scripts and belong to different language families: Afroasiatic, Dravidian and Sino-Tibetan, respectively. Our dataset includes questions in the four languages and their associated SPARQL queries.

| CFQ field | Content |
| --- | --- |
| `questionWithBrackets` | Did ['Murder' Legendre]'s male actor marry [Lillian Lugosi] |
| `questionPatternModEntities` | Did M0 's male actor marry M2 |
| `questionWithMids` | Did m.0h4y854 's male actor marry m.0hpnx3b |
| `sparql` | `SELECT count(*) WHERE { ?x0 ns:film.actor.film/ns:film.performance.character ns:m.0h4y854 . ?x0 ns:people.person.gender ns:m.05zppz . ?x0 ns:people.person.spouse_s/ns:fictional_universe.marriage_of_fictional_characters.spouses ns:m.0hpnx3b . FILTER ( ?x0 != ns:m.0hpnx3b )}` |
| `sparqlPatternModEntities` | `SELECT count(*) WHERE { ?x0 ns:film.actor.film/ns:film.performance.character M0 . ?x0 ns:people.person.gender ns:m.05zppz . ?x0 ns:people.person.spouse_s /ns:fictional_universe.marriage_of_fictional_characters.spouses M2 . FILTER ( ?x0 != M2 )}` |

Table 1: Selected fields in a CFQ entry. `questionWithBrackets` is the full English question with entities surrounded by brackets. `questionPatternModEntities` is the question with entites replaced by placeholders. In `questionWithMids`, the entity codes (Freebase machine IDs; MIDs) are given instead of their labels. `sparql` is the fully executable SPARQL query for the question, and in `sparqlPatternModEntities` the entity codes are replaced by placeholders.

Our contributions are:

- a method to automatically migrate a KBQA dataset to another KB and extend it to diverse languages and domains,

- a benchmark for measuring compositional generalization in SP for KBQA over Wikidata in four typologically diverse languages,

- monolingual experiments with different SP architectures in each of the four languages, demonstrating similar within-language generalization, and

- zero-shot cross-lingual experiments using pretrained multilingual encoders, showing that compositional generalization from English to the other languages fails.

Our code for generating the dataset and for the experiments, as well as the dataset itself and trained models, are publicly available on https://github.com/coastalcph/seq2sparql.

## 2 Limitations of CFQ

CFQ (Compositional Freebase Questions; Keysers et al., 2020) is a dataset for measuring compositional generalization in SP. It targets the task of parsing questions in English into SPARQL queries executable on the Freebase KB (Bollacker et al., 2008).

CFQ contains questions as in Table 1, as well as the following English question (with entities surrounded by brackets):

> "Was [United Artists] founded by [Mr. Fix-it]'s star, founded by [D. W. Griffith], founded by [Mary Pickford], and founded by [The Star Boarder]'s star?"

Parsers trained on CFQ transform these questions into SPARQL queries, which can subsequently be executed against Freebase to answer the original questions (in this case, "Yes").

CFQ uses the Distribution-Based Compositionality Assessment (DBCA) method to generate multiple train-test splits with maximally divergent examples in terms of compounds, while maintaining a low divergence in terms of primitive elements (atoms). In these *maximum compound divergence* (MCD) splits, the test set is constrained to examples containing novel compounds, i.e., new ways of composing the atoms seen during training. For measuring compositional generalizations, named entities in the questions are anonymized so that models cannot simply learn the relationship between entities and properties. CFQ contains 239,357 English question-answer pairs, which encompass 49,320 question patterns and 34,921 SPARQL query patterns. Table 1 shows selected fields of an example in CFQ. In their experiments, Keysers et al. (2020) trained semantic parsers using several architectures on various train-test splits. They demonstrated strong negative correlation between models' accuracy (correctness of the full generated SPARQL query) and compound divergence across a variety of system architectures - all models generalized poorly in the high-divergence settings, highlighting the need to improve compositional generalization in SP.

By the time CFQ was released, Freebase had already been shut down. On that account, to our knowledge, there is no existing SP dataset targeting compositional generalization that is grounded in a currently usable KB, which contains up-to-date information. We therefore migrate the dataset to such a KB, namely Wikidata, in §3.

Moreover, only few studies have evaluated semantic parsers' performance in a multilingual setting, due to the scarcity of multilingual KBQA datasets (Perevalov et al., 2022b). No comparable benchmark exists for languages other than English, and it is therefore not clear whether results are generalizable to other languages. Compositional generalization in typologically distant languages may pose completely different challenges, as these languages may have different ways to compose meaning (Evans and Levinson, 2009). We create such a multilingual dataset in §4, leveraging the multilinguality of Wikidata.

## 3 Migration to Wikidata

Wikidata is widely accepted as the replacement for Freebase. It is actively maintained and represents knowledge in a multitude of languages and domains, and also supports SPARQL. Migrating Freebase queries to Wikidata, however, is not trivial, as there is no established full mapping between the KBs' properties and entities. An obvious alternative to migration would be a replication of the original CFQ generation process but with Wikidata as the KB. Before delving into the details of the migration process, let us motivate the decision not to pursue that option: the grammar used to generate CFQ was not made available to others by Keysers et al. (2020) and is prohibitively too complex to reverse-engineer. Our migration process, on the other hand, is general and can similarly be applied for migrating other datasets from Freebase to Wikidata. Finally, many competitive models with specialized architecture have been developed for CFQ (Guo et al., 2020; Herzig et al., 2021; Gai et al., 2021). Our migrated dataset is formally similar and facilitates their evaluation and the development of new methods.

### 3.1 Property Mapping

As can be seen in Table 1, the WHERE clause in a SPARQL query consists of a list of triples, where the second element in each triple is the property, e.g., `ns:people.person.gender`. CFQ uses 51 unique properties in its SPARQL queries, mostly belonging to the cinematography domain. These Freebase properties cannot be applied directly to Wikidata, which uses different property codes known as P-codes, e.g., `P21`. We therefore need to map the Freebase properties into Wikidata properties.

As a first step in the migration process, we check which Freebase properties used in CFQ have corresponding Wikidata properties. Using a publicly available repository providing a partial mapping between the KBs,[1] we identify 22 out of the 51 Freebase properties in CFQ can be directly mapped to Wikidata properties.[2] The other 29 require further processing:

Fourteen properties are the reverse of other properties, which do not have Wikidata counterparts. For example, `ns:film.director.film` is the reverse of `ns:film.film.directed_by`, and only the latter has Wikidata mapping, `P57`. We resolve the problem by swapping the entities around the property.

The other 15 properties deal with judging whether an entity has a certain quality. In CFQ, `?x1 a ns:film.director` asks whether `?x1` is a director. Wikidata does not contain such unary properties. Therefore, we need to treat these CFQ properties as entities in Wikidata. For example, *director* is `wd:Q2526255`, so we paraphrase the query as `?x1 wdt:P106 wd:Q2526255`, asking whether `?x1`'s *occupation* (`P106`) is director. In addition, we substitute the *art director* property from CFQ with the *composer* property because the former has no equivalent in Wikidata. Finally, we filter out queries with reverse marks over properties, e.g., `?x0 ^ns:people.person.gender M0`, due to incompatibility with the question generation process (§3.2).

After filtering, we remain with 236,304 entries with only fully-mappable properties—98.7% of all entries in CFQ. We additionally make necessary SPARQL syntax modification for Wikidata.[3]

### 3.2 Entity Substitution

A large number of entities in Freebase are absent in Wikidata. For example, neither of the entities in Table 1 exist in Wikidata. Furthermore, unlike the case of properties, to our knowledge, there is no com-prehensive or even partial mapping of Freebase entity IDs (i.e., Freebase machine IDs, MIDs, such as s:m.05zppz) to Wikidata entity IDs (i.e., Q-codes, such as wd:Q6581097). We replicate the grounding process carried out by Keysers et al. (2020), substituting entity placeholders with compatible entities codes by executing the queries against Wikidata:

1. Replacing entity placeholders with SPARQL **variables** (e.g., ?v0), we obtain queries that return sets of compatible candidate entity assignments instead of simply an answer for a given assignment of entities.

2. We add constraints for the entities to be **distinct**, to avoid nonsensical redundancies (e.g., due to conjunction of identical clauses).

3. Special entities, representing **nationalities and genders**, are regarded as part of the question patterns in CFQ (and are not replaced with placeholders). Before running the queries, we thus replace all such entities with corresponding Wikidata Q-codes (instead of variables).

4. We **execute** the queries against the Wikidata query service[4] to get the satisfying assignments of entity combinations, with which we replace the placeholders in sparqlPatternModEntities fields.

---

[1] https://www.wikidata.org/wiki/Wikidata:WikiProject_Freebase/Mapping

[2] While some Freebase properties have multiple corresponding Wikidata properties, we consider a property mappable as long as it has at least one mapping.

[3] CFQ uses SELECT count(*) WHERE to query yes/no questions, but this syntax is not supported by Wikidata. We replace it with ASK WHERE, intended for boolean queries.

| Lang. | MCWQ field | Content |
|---|---|---|
| En | questionWithBrackets | Did [Lohengrin] 's male actor marry [Margarete Joswig] |
|  | questionPatternModEntities | Did M0 's male actor marry M2 |
| He | questionWithBrackets | האם השחקן הגברי של [לוהנגרין] התחתן עם [מרגרט יוסוויג] |
|  | questionPatternModEntities | האם השחקן הגברי של M0 התחתן עם M2 |
| Kn | questionWithBrackets | [ಲೋಹೆಂಗ್ರಿನ್] ಅವರ ಪುರುಷ ನಟ ವಿವಾಹವಾದರು [ಮಾರ್ಗರೆಟ್ ಜೋಸ್ವಿಗ್] |
|  | questionPatternModEntities | M0 ನ ಪುರುಷ ನಟ M2 ಅನ್ನು ಮದುವೆಯಾಗಿದ್ದಾರೆಯೆ |
| Zh | questionWithBrackets | [Lohengrin]的男演员嫁给了[Margarete Joswig]吗 |
|  | questionPatternModEntities | M0的男演员和M2结婚吗 |
|  | sparql | ASK WHERE { ?x0 wdt:P453 wd:Q50807639 . ?x0 wdt:P21 wd:Q6581097 . ?x0 wdt:P26 wd:Q1560129 . FILTER ( ?x0 != wd:Q1560129 )} |
|  | sparqlPatternModEntities | ASK WHERE { ?x0 wdt:P453 M0 . ?x0 wdt:P21 wd:Q6581097 . ?x0 wdt:P26 M2 . FILTER ( ?x0 != M2 )} |
|  | recursionDepth | 20 |
|  | expectedResponse | True |

Table 2: The MCWQ example from Figure 1. The English question is generated from the CFQ entry in Table 1 by the migration process described in §3.3, and the questions in the other languages are automatically translated (§4.1). The questionWithBrackets, questionPatternModEntities, sparql and sparqlPatternModEntities fields are analogous to the CFQ ones. recursionDepth (which quantifies the question complexity) and expectedResponse (which is the answer returned upon execution of the query) are copied from the CFQ entry.

[4] https://query.wikidata.org/

5. Finally, we insert the Q-codes into the English **questions** in the questionWithMids field and the corresponding entity labels into the questionWithBrackets to obtain the English questions for our dataset.

Along this process, 52.5% of the queries have at least one satisfying assignment. The resulting question-query pairs constitute our English dataset. They maintain the SPARQL patterns in CFQ, but the queries are all executable on Wikidata.

We obtain 124,187 question-query pairs, of which 67,523 are yes/no questions and 56,664 are wh- questions. The expected responses of yes/no questions in this set are all "yes" due to our entity assignment process. To make MCWQ comparable to CFQ, which has both positive and negative answers, we sample alternative queries by replacing entities with ones from other queries whose preceding predicates are the same. Our negtive sampling results in 30,418 questions with "no" answers.

### 3.3 Migration Example

Consider the SPARQL pattern from Table 1:

```
SELECT count(*) WHERE { ?x0 ns:film.actor.
    film/ns:film.performance.character M0 . ?
    x0 ns:people.person.gender ns:m.05zppz .
    ?x0 ns:people.person.spouse_s/ns:
    fictional_universe.
    marriage_of_fictional_characters.spouses
    M2 . FILTER ( ?x0 != M2 )}
```

We replace the properties and special entities (here the gender *male*: ns:m.05zppz → wd:Q6581097):

```
SELECT count(*) WHERE {?x0 wdt:P453 M0 .?x0
    wdt:P21 wd:Q6581097 . ?x0 wdt:P26 M2 .
    FILTER ( ?x0 != M2 )}
```

Then we replace placeholders (e.g., `M0`) with variables and add constraints for getting only one assignment (which is enough for our purposes) with distinct entities. The resulting query is:

```
SELECT ?v0 ?v1 WHERE {?x0 wdt:P453 ?v0 . ?x0
    wdt:P21 wd:Q6581097 . ?x0 wdt:P26 ?v1 .
    FILTER ( ?x0 != ?v1 ) . FILTER ( ?v0 != ?
    v1 )} LIMIT 1
```

We execute the query and get `wd:Q50807639` (Lohengrin) and `wd:Q1560129` (Margarete Joswig) as satisfying answers for `v0` and `v1` respectively. Note that these are different from the entities in the original question ('Murder' Legendre and Lillian Lugosi)—in general, there is no guarantee that the same entities from CFQ will be preserved in our dataset. Then we put back these answers into the query, and make necessary SPARQL syntax modification for Wikidata. The final query for this entry is:

```
ASK WHERE {?x0 wdt:P453 wd:Q50807639 . ?x0 wdt
    :P21 wd:Q6581097 . ?x0 wdt:P26 wd:
    Q1560129 . FILTER ( ?x0 != wd:Q1560129 )}
```

As for the English question, we map the Freebase entities in the `questionWithMids` field with the labels of the obtained Wikidata entities. Therefore, the English question resulting from this process is:

> Did [Lohengrin] 's male actor marry [Margarete Joswig]?

### 3.4 Dataset Statistics

We compare the statistics of MCWQ with CFQ in Table 3. MCWQ has 29,312 unique question patterns (mod entities, verbs, etcs), i.e., 23.6% of questions cover all question patterns, compared to 20.6% in CFQ. Furthermore, MCWQ has 86,353 unique query patterns (mod entities), resulting in 69.5% of instances covering all SPARQL patterns, 18% higher than CFQ. Our dataset thus poses a greater challenge for compositional SP, and exhibits less redundancy in terms of duplicate query patterns. It is worth noting that less unique query percentage in MCWQ than CFQ results from the loss during swapping the entities in §3.1.

To be compositionally challenging, Keysers et al. (2020) generated the MCD splits to have high compound divergence while maintaining low atom divergence. As atoms in MCWQ are mapped from CFQ while leaving the compositional structure intact, we derive train-test splits of our dataset by inducing the train-test splits from CFQ on the corresponding subset of instances in our dataset.

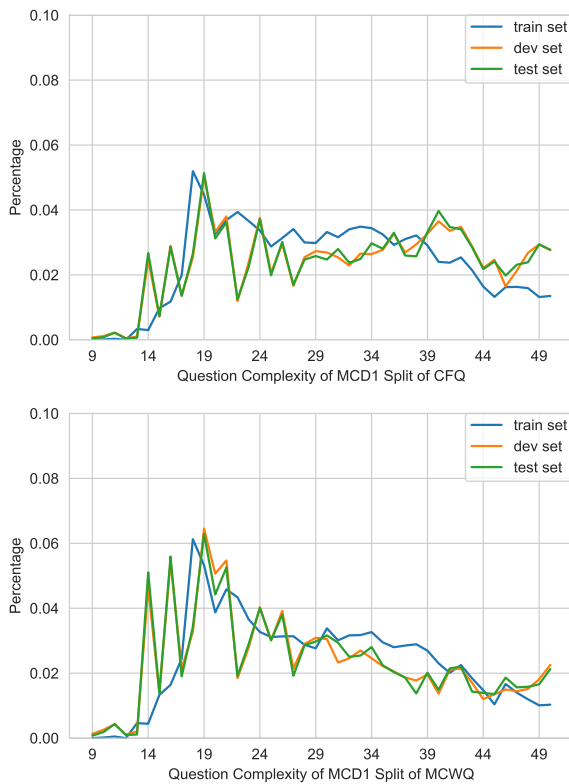

Figure 2: Complexity distribution of the MCD$_1$ split of CFQ (above) and MCWQ (below).

The complexity of questions in CFQ is measured by recursion depth and reflects the number of rule applications used to generate a question, which encompasses grammar, knowledge, inference and resolution rules. While each question's complexity in MCWQ is the same as the corresponding CFQ question's, some cannot be migrated (see §3.1 and §3.2). To verify the compound divergence is not affected, we compare the question complexity distribution of the two datasets in one of the three compositional splits (MCD1) in Figure 2. The training, development and test sets of the split in CFQ and MCWQ follow a similar trend in general. The fluctuation in the complexity of questions in the MCWQ splits reflects the dataset's full distribution—see Figure 3.

Stemming from its entities and properties, CFQ questions are limited to the domain of movies. The entities in MCWQ, however, can in principle come from any domain, owing to our flexible entity replacing method. Though MCWQ's properties are still a subset of those used in CFQ, they are primarily in the movies domain. We also observe a few questions

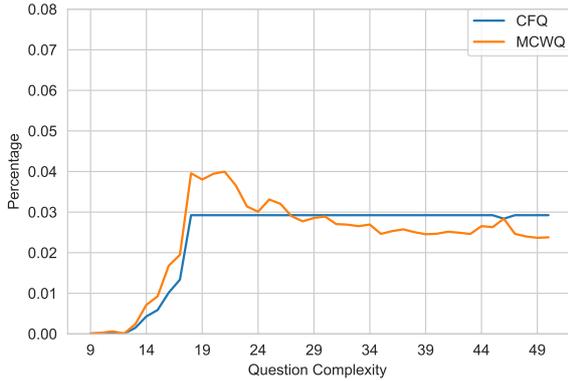

Figure 3: Complexity distribution of MCWQ, measured by recursion depth, compared to CFQ.

|  | CFQ | MCWQ |
| --- | --- | --- |
| Unique questions | 239,357 | 124,187 |
| Questions patterns | 49,320 (20.6%) | 29,312 (23.6%) |
| Unique queries | 228,149 (95.3%) | 101,856 (82%) |
| Query patterns | 123,262 (51.5%) | 86,353 (69.5%) |
| Yes/no questions | 130,571 (54.6%) | 67,523 (54.4%) |
| Wh- questions | 108,786 (45.5%) | 56,664 (45.6%) |

Table 3: Dataset statistics comparison for MCWQ and CFQ. Percentages are relative to all unique questions. Questions patterns refer to mod entities, verbs, etc. while query patterns refer to mod entities only.

from literature, politics, and history in MCWQ.

## 4 Generating Multilingual Questions

To create a typologically diverse dataset, starting from our English dataset (an Indo-European language using the Latin script), we use machine translation to three other languages from different families (Afroasiatic, Dravidian and Sino-Tibetan), which use different scripts: Hebrew, Kannada and Chinese (§4.1). For a comparison to machine translation and a more realistic evaluation with regards to compositional SP, we manually translate a subset of the test sets of the three MCD splits (§4.2) and evaluate the machine translation quality (§4.3).

### 4.1 Generating Translations

Both question patterns and bracketed questions are translated separately with Google Cloud Translation[5] from English.[6] SPARQL queries remain unchanged, as both property and entity IDs are language-independent in Wikidata, which contains labels in different languages for each. Table 2 shows an example for a question in our dataset (which is generated from the same question as the CFQ instance from Table 1), as well as the resulting translations.

As an additional technical necessity, we add a question mark to the end of each question before translation (as the original dataset does not include question marks) and remove trailing question marks from the translated question before including it in our dataset. We find this step to be essential for translation quality.

### 4.2 Gold Test Set

CFQ and other datasets for evaluating compositional generalization (Lake and Baroni, 2018; Kim and Linzen, 2020) are generated from grammars. However, It has not been investigated how well models trained on them generalize to human questions. As a step towards that goal, we evaluate whether models trained with automatically generated and translated questions can generalize to high-quality human-translated questions. For that purpose, we obtain the intersection of the test sets of the MCD splits (1,860 entries), and sample two translated questions with yes/no questions and two with wh- questions for each complexity level (if available). This sample, termed *test-intersection-MT*, has 155 entries in total. The authors (one native speaker for each language) manually translate the English questions into Hebrew, Kannada and Chinese. We term the resulting dataset *test-intersection-gold*.

### 4.3 Translation Quality

We compute the BLEU (Papineni et al., 2002) scores of *test-intersection-MT* against *test-intersection-gold* using SacreBLEU (Post, 2018), resulting in 87.4, 76.6 and 82.8 for Hebrew, Kannada and Chinese, respectively. This indicates high quality of the machine translation outputs.

Additionally, one author for each language manually assesses translation quality for one sampled question from each complexity level from the full dataset (40 in total). We rate the translations on a scale of 1–5 for fluency and for meaning preservation, with 1 being poor, and 5 being optimal. Despite occasional translation issues, mostly attributed to lexical choice or morphological agreement, we confirm that the translations are of high quality. Across languages, over 80% of examples score 3 or higher in fluency and meaning preservation. The average meaning preservation scores for Hebrew, Kannada and Chinese are 4.4, 3.9 and 4.0. For fluency, they are 3.6, 3.9 and 4.4.

---

[5]https://cloud.google.com/translate
[6]We attempted to translate bracketed questions and subsequently replace the bracketed entities with placeholders as question patterns. In preliminary experiments, we found that separate translation of question patterns is of higher translation quality. Therefore, we choose to translate question patterns and bracketed questions individually.

As a control, one of the authors (a native English speaker) evaluates English fluency for the same sample of 40 questions. Only 62% of patterns were rated 3 or above. While all English questions are grammatical, many suffer from poor fluency, tracing back to their automatic generation using rules. Some translations are rated higher in terms of fluency, mainly due to annotator leniency (focusing on disfluencies that might result from translation) and paraphrasing of unnatural constructions by the MT system (especially for lower complexities).

## 5 Experiments

While specialized architectures have been achieved state-of-the-art results on CFQ (Guo et al., 2020, 2021; Gai et al., 2021), these approaches are English- or Freebase-specific. We therefore experiment with sequence-to-sequence (seq2seq) models, among which T5 (Raffel et al., 2020) has been shown to perform best on CFQ (Herzig et al., 2021). We evaluate these models for each lanuage separately (§5.1), and subsequently evaluate their cross-lingual compositional generalization (§5.2).

### 5.1 Monolingual Experiments

We evaluate six models' monolingual parsing performance on the three MCD splits and a random split of MCWQ. As done by Keysers et al. (2020), entities are masked during training, except those that are part of the question patterns (genders and nationalities).

We experiment with two seq2seq architectures on MCWQ for each language, with the same hyperparameters tuned by Keysers et al. (2020) on the CFQ random split: LSTM (Hochreiter and Schmidhuber, 1997) with attention mechanism (Bahdanau et al., 2015) and Evolved Transformer (So et al., 2019), both implemented using Tensor2Tensor (Vaswani et al., 2018). Separate models are trained and evaluated per language, with randomly initialized (not pretrained) encoders. We train a model for each of the three MCD splits plus a random split for each language.

We also experiment with pretrained language models (PLMs), to assess whether *multilingual* PLMs, mBERT (Devlin et al., 2019) and mT5 (Xue et al., 2020), are as effective for monolingual compositional generalization as an English-only PLM using the Transformers library (Wolf et al., 2020).

For mBERT, we fine-tune a `multi_cased_L-12_H-768_A-12` encoder and a randomly initialized decoder of the same architecture. We train for 100 epochs with patience of 25, batch size of 128, and learning rate of $5 \times 10^{-5}$ with a linear decay.

For T5, we fine-tune `T5-base` on MCWQ English, and `mT5-small` and `mT5-base` on each language separately. We use the default hyperparameter settings except trying two learning rates, $5e^{-4}$ and $3e^{-5}$ (see results below). SPARQL queries are preprocessed using reversible intermediate representations (RIR), previously shown (Herzig et al., 2021) to facilitate compositional generalization for T5. We fine-tune all models for 50K steps.

We use six Titan RTX GPUs for training, with batch size of 36 for `T5-base`, 24 for `mT5-small` and 12 for `mT5-base`. We use two random seeds for `T5-base`. It takes 384 hours to finish a round of `mT5-small` experiments, 120 hours for `T5-base` and 592 hours for `mT5-base`.

In addition to exact-match accuracy, we report the BLEU scores of the predictions computed with SacreBLEU, as a large portion of the generated queries is partially (but not fully) correct.

**Results** The results are shown in Table 4. While models generalize almost perfectly in the random split for all four languages, the MCD splits are much harder, with the highest mean accuracies of 38.3%, 33.2%, 32.1% and 36.3% for English, Hebrew, Kannada and Chinese, respectively. For comparison, on CFQ, `T5-base+RIR` has an accuracy of 60.8% on $\text{MCD}_{mean}$ (Herzig et al., 2021). One reason for this decrease in performance is the smaller training data: the MCWQ dataset has 52.5% the size of CFQ. Furthermore, MCWQ has less redundancy than CFQ in terms of duplicate questions and SPARQL patterns, rendering models' potential strategy of simply memorizing patterns less effective.

Contrary to expectation, `mT5-base` does not outperform `mT5-small`. During training, we found `mT5-base` reached minimum loss early (after 1k steps). By changing the learning rate from the default $3e^{-5}$ to $5e^{-4}$, we seem to have overcome the local minimum. Training `mT5-small` with learning rate $5e^{-4}$ also renders better performance. Furthermore, the batch size we use for `mT5-base` may not be optimal, but we could not experiment with larger batch sizes due to resource limitations.

| Exact Match (%) | MCD₁ | | | | MCD₂ | | | | MCD₃ | | | | MCD_mean | | | | Random | | | |
|---|---|---|---|---|---|---|---|---|---|---|---|---|---|---|---|---|---|---|---|---|
| | En | He | Kn | Zh | En | He | Kn | Zh | En | He | Kn | Zh | En | He | Kn | Zh | En | He | Kn | Zh |
| LSTM+Attention | 38.2 | 29.3 | 27.1 | 26.1 | 6.3 | 5.6 | 9.9 | 7.5 | 13.6 | 11.5 | 15.7 | 15.1 | 19.4 | 15.5 | 17.6 | 16.2 | 96.6 | 80.8 | 88.7 | 86.8 |
| E. Transformer | 53.3 | 35 | 30.7 | 31 | 16.5 | 8.7 | 11.9 | 10.2 | 18.2 | 13 | 18.1 | 15.5 | 29.3 | 18.9 | 20.2 | 18.9 | 99 | 90.4 | 93.7 | 92.2 |
| mBERT | 49.5 | 38.7 | 34.4 | 35.6 | 13.4 | 11.4 | 12.3 | 15.1 | 17 | 18 | 18.1 | 19.4 | 26.6 | 22.7 | 21.6 | 23.4 | 98.7 | 91 | 95.1 | **93.3** |
| T5-base+RIR | 57.4 | - | - | - | 14.6 | - | - | - | 12.3 | - | - | - | 28.1 | - | - | - | 98.5 | - | - | - |
| mT5-small+RIR | **77.6** | 57.8 | 55 | **52.8** | 13 | 12.6 | 8.2 | 21.1 | **24.3** | 17.5 | **31.4** | 34.9 | **38.3** | 29.3 | 31.5 | **36.3** | 98.6 | 90 | 93.8 | 91.8 |
| mT5-base+RIR | 55.5 | **59.5** | 49.1 | 30.2 | **27.7** | **16.6** | **16.6** | **23** | 18.2 | **23.4** | 30.5 | **35.6** | 33.8 | **33.2** | **32.1** | 29.6 | **99.1** | **90.6** | **94.2** | 92.2 |

Table 4: Monolingual evaluation: exach match accuracies on MCWQ. MCD_mean is the mean accuracy of all three MCD splits. Random represents a random split of MCWQ. This is an upper bound on the performance shown only for comparison. As SPARQL BLEU scores are highly correlated with accuracies in this experiment, we only show the latter here.

Comparing the performance across languages, `mT5-base` performs best on Hebrew and Kannada on average, while `mT5-small` has the best performance on English and Chinese. Due to resource limitations, we were not able to look deeper into the effect of hyperparameters or evaluate larger models. However, our experiments show that while multilingual compositional generalization is challenging for seq2seq semantic parsers, within-language generalization is comparable between languages. Nonetheless, English is always the easiest (at least marginally). A potential cause is that most semantic query languages were initially designed to represent and retrieve data stored in English databases, and thus have a bias towards English. Consequently, SPARQL syntax is closer to English than Hebrew, Kannada and Chinese. While translation errors might have an effect as well, we have seen in §4.3 that translation quality is high.

To investigate further, we plot the complexity distribution of true predictions (exactly matching the gold SPARQL) per language by the two best systems in Figure 4. We witness a near-linear performance decay from complexity level 19. We find that `mT5-base` is better than `mT5-small` on lower complexity despite the latter's superior overall performance. Interestingly, translated questions seem to make the parsers generalize better at higher complexity, as shown in the figure. For `mT5-small`, the three non-English models successfully parse more questions within the complexity range 46-50 than English, for `mT5-base` 44-50. As is discussed in §4.3, machine-translated questions tend to have higher fluency than English questions; we conjecture such a smoothing method helps the parser to understand and learn from higher complexity questions.

### 5.2 Zero-shot Cross-lingual Parsing

Zero-shot cross-lingual SP has witnessed new advances with the development of PLMs (Shao et al.,

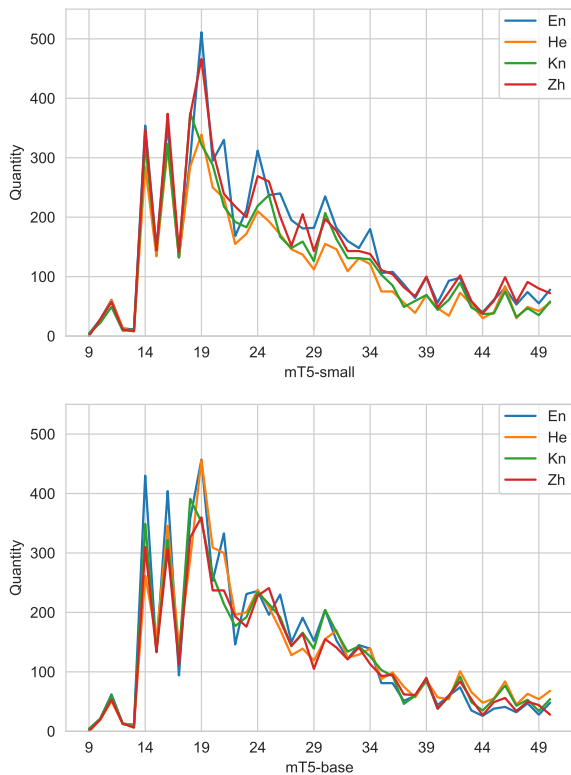

Figure 4: Two mT5 models' number of correct predictions summing over the three MCD splits in monolingual experiments, plotted by complexity level. Each line represents a language. While `mT5-small` generalizes better overall, `mT5-base` is better in lower complexities (which require less compositional generalization).

2020; Sherborne and Lapata, 2022). Since translating datasets and training KBQA systems is expensive, it is beneficial to leverage multilingual PLMs, fine-tuned on English data, for generating SPARQL queries over Wikidata given natural language questions in different languages. While compositional generalization is difficult even in a monolingual setting, it is interesting to investigate whether multilingual PLMs can transfer in cross-lingual SP over

|  | MCD$_{mean}$ | | | | Random | | | |
|---|---|---|---|---|---|---|---|---|
| **SPARQL BLEU** | En | He | Kn | Zh | En | He | Kn | Zh |
| mT5-small+RIR | 87.5 | 53.8 | 53.2 | 59 | 99.9 | 60.4 | 59.9 | 63.8 |
| mT5-base+RIR | 86.4 | 46.4 | 46 | 52.7 | 99.9 | 63.2 | 63.5 | 70.6 |
| **Exact Match (%)** | | | | | | | | |
| mT5-small+RIR | 38.3 | 0.2 | 0.3 | 0.2 | 98.6 | 0.5 | 0.4 | 1.1 |
| mT5-base+RIR | 33.8 | 0.4 | 0.7 | 1.5 | 99.1 | 1.1 | 0.9 | 7.2 |

Table 5: Mean BLEU scores and exact match accuracies on the three MCD splits and on a random split in zero-shot cross-lingual transfer experiments on MCWQ. The grey texts represent the models' monolingual performance on English, given for reference (the exact match accuracies are copied from Table 4). The black texts indicate the zero-shot cross-lingual transfer performances on Hebrew, Kannada and Chinese of a model trained on English. While the scores for individual MCD splits are omitted for brevity, in all three MCD splits, the accuracies are below 1% (except on MCD$_2$ Chinese, being 4%).

|  | *test-intersection-MT* | | | | *test-intersection-gold* | | | |
|---|---|---|---|---|---|---|---|---|
| **SPARQL BLEU** | En | He | Kn | Zh | En | He | Kn | Zh |
| mT5-small+RIR | 86.1 | 82.5 | 78.9 | 85.1 | - | 81.8 | 77.7 | 86 |
| mT5-base+RIR | 85.5 | 83.7 | 81.8 | 83.2 | - | 83.8 | 80.9 | 83.8 |
| **Exact Match (%)** | | | | | | | | |
| mT5-small+RIR | 45.6 | 35.7 | 32.7 | 38.5 | - | 35.9 | 28.2 | 39.8 |
| mT5-base+RIR | 40.4 | 41.9 | 40.2 | 38.7 | - | 41.1 | 34 | 38.9 |

Table 6: Mean BLEU scores and accuracies of monolingual models (§5.1) on *test-intersection-MT* and *test-intersection-gold*. The numbers are averaged over the accuracies of the predictions from the monolingual models trained on three MCD splits. Overall, there is no substantial difference between the performances on the two intersection sets, demonstrating the reliability of evaluating on machine translated data in this case.

**Wikidata.** Simple seq2seq T5/mT5 models perform reasonably well (> 30% accuracy) on monolingual SP on some splits (see §5.1). We investigate whether the learned multilingual representations of such models enable compositional generalization even without target language training. We use mT5-small+RIR and mT5-base+RIR, the best two models trained and evaluated on English from previous experiments, to predict on the other languages.

**Results** The results are shown in Table 5. Both BLEU and exact match accuracy of the predicted SPARQL queries drop drastically when the model is evaluated on Hebrew, Kannada and Chinese. mT5-small+RIR achieves 38.3% accuracy on MCD$_{mean}$ English, but less than 0.3% in zero-shot parsing on three non-English languages.

Even putting aside compositionality evaluation, as seen in the random split, the exact match accuracy in the zero-shot cross-lingual setting is still low. The relatively high BLEU scores can be attributed to the small overall vocabulary used in SPARQL queries. Interestingly, while mT5-base+RIR on MCD$_{mean}$ English does not outperform mT5-small+RIR, it yields better performance in the zero-shot setting. For Hebrew, Kannada and Chinese, the accuracies are 0.2%, 0.4% and 1.3% higher. For mT5-base, Chinese is slightly easier than Kannada and Hebrew to parse in the zero-shot setting, outperforming 1.1% and 0.8%.

To conclude, zero-shot cross-lingual transfer from English to Hebrew, Kannada and Chinese fails to generate valid queries in MCWQ. A potential cause for such unsuccessful transfer is that all four languages in MCWQ belong to different language families and have low linguistic similarities. It remains to be investigated whether such cross-lingual transfer will be more effective on related languages, such as from English to German (Lin et al., 2019).

## 6 Analysis

### 6.1 Evaluation with Gold Translation

Most existing compositional generalization datasets focus on SP (Lake and Baroni, 2018; Kim and Linzen, 2020; Keysers et al., 2020). These datasets are composed either with artificial language or in English using grammar rules. With *test-intersection-gold* proposed in §4.2, we investigate whether models can generalize from a synthetic automatically translated dataset to a manually translated dataset.

We use the monolingual models trained on three MCD splits to parse *test-intersection-gold*. In Table 6, we present the mean BLEU scores and exact match accuracy of the predicted SPARQL queries. There is no substantial difference between the performances on the two intersection sets, except for Kannada, which has a 4% accuracy drop on average. These results testify that MCWQ has sufficiently high translation quality and that models trained with such synthetic data can be used to generalize to high-quality manually-translated questions.

### 6.2 Categorizing Errors

In an empirical analysis, we categorize typical prediction errors on *test-intersection-gold* and *test-intersection-MT* into six types: missing property, extra property, wrong property (where the two property sets have the same numbers of properties, but the elements do not match), missing entity, extra entity and wrong entity (again, same number of entities but different entity sets). We plot the mean number of errors per category, as well as the number of predictions with multiple errors, in Figure 5 for monolingual `mT5-small` models. Overall, model predictions tend to have more missing properties and entities than extra ones. Different languages, however, vary in error types. For example, on Hebrew, models make more missing property/entity errors than other languages; but on Kannada they make more extra property/entity errors than the others. About 70 out of the 155 examples contain multiple errors for all languages, with Kannada being slightly more.

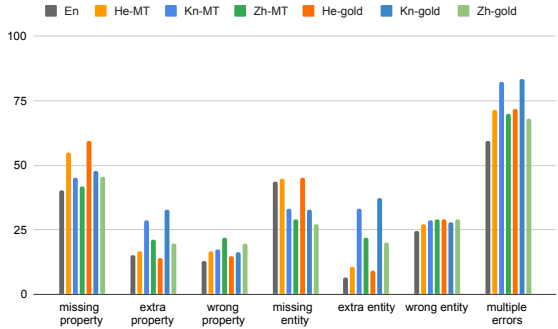

Figure 5: Number of errors per category in different SPARQL predictions on *test-intersection-MT* and *test-intersection-gold*, averaged across monolingual `mT5-small+RIR` models trained on the three MCD splits. The total number of items in each test set is 155.

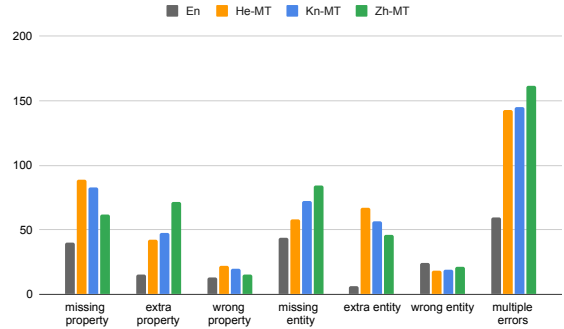

Figure 6: Number of errors per category in different zero-shot cross-lingual SPARQL predictions on *test-intersection-MT*, averaged across `mT5-small+RIR` models trained on the three MCD splits in English. Additionally, mean error counts on the English set are given for comparison. The total number of items in each test set is 155.

Comparing errors on *test-intersection-gold* and *test-intersection-MT*, we find missing properties are more common in *gold* for all languages. For Hebrew and Kannada, extra properties and entities are also more common in *gold*. However, for Chinese, these and missing entities are less common in *gold* compared to *MT*.

In Figure 6 we plot the error statistics for zero-shot cross-lingual transfer using `mT5-small` models. We can see there are drastically more error occurrences. For both missing and extra property/entity, the numbers are about double those from monolingual experiments. The number of wrong property/entity errors remain similar, due to the difficulty of even predicting a set of the correct size in this setting. For all three target languages, nearly all predictions contain multiple errors. The statistics indicate the variety

| Question | Was M0 written by and directed by M1 , M2 , and M3 |
|---|---|
| Gold | `ASK WHERE { M0 wdt:P57 M1 . M0 wdt:P57 M2 . M0 wdt:P57 M3 . M0 wdt:P58 M1 . M0 wdt:P58 M2 . M0 wdt:P58 M3 }` |
| Inferred | `ASK WHERE { M0 wdt:P57 M1 . M1 wdt:P57 M2 . M0 wdt:P58 M3 }` |

Figure 7: Example for an error reflecting incorrect predicate-argument structure. `wdt:P57` is *director* and `wdt:P58` is *screenwriter*. Incorrect triples are shown in red and missed triples in blue.

and pervasiveness of errors.

### 6.3 Other Observations

We also find that comparatively, parsers perform well on short questions on all four languages. This is expected as the compositionality of these questions is inherently low. On languages other than English, the models perform well when the translations are faithful. On occasions when they are less faithful or fluent but still generate correct queries, we hypothesize that translation acts as data regularizers, especially at higher complexities, as demonstrated in Figure 4.

Among wrong entity errors, the most common cause across languages is the shuffling of entity placeholders. In the example shown in Figure 7, we see that the model generates `M1 wdt:P57 M2` instead of `M0 wdt:P57 M2`, which indicates incorrect predicate-argument structure interpretation.

## 7 Related Work

**Compositional Generalization** Compositional generalization has witnessed great developments in recent years. SCAN (Lake and Baroni, 2018), a synthetic dataset consisting of natural language and command pairs, is an early dataset designed to systematically evaluate neural networks' generalization ability. CFQ and COGS are two more realistic benchmarks following SCAN. There are various approaches developed to enhance compositional generalization, for example, by using hierarchical poset decoding (Guo et al., 2020), combining relevant queries (Das et al., 2021) using span representation (Herzig and Berant, 2021) and graph encoding (Gai et al., 2021). In addition to pure language, the evaluation of compositional generalization has been expanded to image captioning and situated language understanding (Nikolaus et al., 2019; Ruis et al., 2020). Multilingual and cross-lingual compositional generalization is an important and challenging field to which our paper aims to bring researchers' attention.

**Knowledge Base Question Answering** Comparing to machine reading comprehension (Rajpurkar et al., 2016; Joshi et al., 2017; Shao et al., 2018; Dua et al., 2019; d'Hoffschmidt et al., 2020), KBQA is less diverse in terms of datasets. Datasets such as WebQuestions (Berant et al., 2013), SimpleQuestions (Bordes et al., 2015), ComplexWebQuestions (Talmor and Berant, 2018), FreebaseQA (Jiang et al., 2019), GrailQA (Gu et al., 2021), CFQ and *CFQ (Tsarkov et al., 2021) were proposed on Freebase, a now-discontinued KB. SimpleQuestions2Wikidata (Diefenbach et al., 2017) and ComplexSequentialQuestions (Saha et al., 2018) are based on Wikidata, but like most others, they are monolingual English datasets. Related to our work is RuBQ (Korablinov and Braslavski, 2020; Rybin et al., 2021), an English-Russian dataset for KBQA over Wikidata. While the dataset is bilingual, it uses crowdsourced questions and is not designed for compositionality analysis. Recently, Thorne et al. (2021) proposed WIKINLDB, a Wikidata-based English KBQA dataset, focusing on scalability rather than compositionality. Other related datasets include QALM (Kaffee et al., 2019), a dataset for multilingual question answering over a set of different popular knowledge graphs, intended to help determine the multilinguality of those knowledge graphs. Similarly, QALD-9 (Ngomo, 2018) and QALD-9-plus (Perevalov et al., 2022a) support the development of multilingual question answering systems, tied to DBpedia and Wikidata, respectively. The goal of both datasets is to expand QA systems to more languages rather than improving compositionality. KQA Pro (Cao et al., 2022), a concurrent work to us, is an English KBQA dataset over Wikidata with a focus on compositional reasoning.

Wikidata has been leveraged across many NLP tasks such as coreference resolution (Aralikatte et al., 2019), frame-semantic parsing (Sas et al., 2020), entity linking (Kannan Ravi et al., 2021) and named entity recognition (Nie et al., 2021). As for KBQA, the full potential of Wikidata is yet to be explored.

**Multilingual and Cross-lingual Modelling** Benchmarks such as XGLUE (Liang et al., 2020) and XTREME (Hu et al., 2020) focus on multilingual classification and generation tasks. Cross-lingual learning has been studied across multiple fields, such as sentiment analysis (Abdalla and Hirst, 2017), document classification (Dong and de Melo, 2019), POS tagging (Kim et al., 2017) and syntactic parsing (Rasooli and Collins, 2017). In recent years, multilingual PLMs have been a primary tool for extending NLP applications to low-resource languages, as these models ameliorate the need to train individual models for each language, for which less data may be available. Several works have attempted to explore the limitations of such models in terms of practical usability for low-resource languages (Wu and Dredze, 2020), and also the underlying elements that make cross-lingual transfer learning viable (Dufter and Schütze, 2020). Beyond these PLMs, other works focus on improving cross-lingual learning by making particular changes to the encoder-decoder architecture, such as adding adapters to attune to specific information (Artetxe et al., 2020b; Pfeiffer et al., 2020).

For cross-lingual SP, Sherborne and Lapata (2022) explored zero-shot SP by aligning latent representations. Zero-shot cross-lingual SP has also been studied in dialogue modelling (Nicosia et al., 2021). Yang et al. (2021) present augmentation methods for Discourse Representation Theory (Liu et al., 2021b). Oepen et al. (2020) explore cross-framework and cross-lingual SP for meaning representations. To the best of our knowledge, our work is the first on studying cross-lingual transfer learning in KBQA.

## 8 Limitations

MCWQ is based on CFQ, a rule-base generated dataset, and hence the inherited unnaturalness in question-query pairs of high complexity. Secondly, we use machine translation to make MCWQ multilingual. Although this is the dominant approach for generating multilingual datasets (Ruder et al., 2021) and we have provided evidences that MCWQ has reasonable translation accuracy and fluency with human evaluation and comparative experiments in §4.3 and §5.1, machine translation would nevertheless create substandard translation artifacts (Artetxe et al., 2020a). One alternative way is to write rules for template translation. The amount of work can possibly be reduced by refering to a recent work (Goodwin et al., 2021) in which English rules are provided for syntactic dependency parsing on CFQ's question fields.

| MCWQ mT5-base+RIR      |                  |
|------------------------|------------------|
| **Information**        | **Unit**         |
| 1. Model publicly available? | Yes        |
| 2. Time to train final model | 592 hours  |
| 3. Time for all experiments | 1315 hours  |
| 4. Energy consumption  | 2209.2 kWh       |
| 5. Location for computations | Denmark    |
| 6. Energy mix at location | 191 gCO2eq/ kWh |
| 7. CO2eq for final model | 189.96 kg      |
| 8. CO2eq for all experiments | 421.96 kg  |

Table 7: Climate performance model card for `mT5-base+RIR` fine-tuned on all splits and languages.

Furthermore, the assumption that an English KB is a "canonical" conceptualization is unjustified, as speakers of other languages may know and care about other entities and relationships (Liu et al., 2021a; Hershcovich et al., 2022a). Therefore, future work must create multilingual SP datasets by sourcing questions from native speakers rather than translating them.

## 9 Conclusion

The field of KBQA has been saturated with work on English, due to both the inherent challenges of translating datasets and the reliance on English-only DBs. In this work, we presented a method for migrating the existing CFQ dataset to Wikidata and created a challenging multilingual dataset, MCWQ, targeting compositional generalization in multilingual and cross-lingual SP. In our experiments, we observe that pretrained multilingual language models struggle to transfer and generalize compositionally across languages. Our dataset will facilitate building robust multilingual semantic parsers by serving as a benchmark for evaluation of cross-lingual compositional generalization.

## 10 Environmental Impact

Following the climate-aware practice proposed by Hershcovich et al. (2022b), we present a climate performance model card in Table 7. "Time to train final model" is the sum over splits and languages for `mT5-base+RIR`, while "Time for all experiments" also includes the experiments with the English-only `T5-base+RIR` across all splits. While the work does not have direct positive environmental impact, better understanding of compositional generalization, resulting from our work, will facilitate more efficient modeling and therefore reduce emissions in the long term.

## 11 Acknowledgments

The authors thank Anders Søgaard and Miryam de Lhoneux for their comments and suggestions, as well as the TACL editors and several rounds of reviewers for their constructive evaluation. This project has received funding from the European Union's Horizon 2020 research and innovation programme under the Marie Skłodowska-Curie grant agreement No 801199 (*Heather Lent*).